\documentclass[journal]{IEEEtran}
\ifCLASSINFOpdf
\else
\fi

\usepackage{times}
\usepackage{epsfig}
\usepackage{textpos}
\usepackage{graphicx}
\usepackage{amsmath}
\usepackage{amssymb}
\usepackage{bm}
\usepackage{booktabs}
\usepackage{cite}
\usepackage{subfigure}

\begin{document}
%
\title{Attention-guided Chained Context Aggregation\\for Semantic Segmentation}
%
%
%


\author{Quan Tang,
        Fagui Liu,~\IEEEmembership{Member,~IEEE,}
        Tong Zhang,~\IEEEmembership{Member,~IEEE,}
        Jun Jiang,
        and Yu Zhang
}

%
%

\markboth{Journal of \LaTeX\ Class Files,~Vol.~14, No.~8, August~2015}%
{Shell \MakeLowercase{\textit{et al.}}: Bare Demo of IEEEtran.cls for IEEE Journals}
%



\maketitle

\begin{abstract}
The way features propagate in Fully Convolutional Networks is of momentous importance to capture multi-scale contexts for obtaining precise segmentation masks. This paper proposes a novel series-parallel hybrid paradigm called the Chained Context Aggregation Module (CAM) to diversify feature propagation. CAM gains features of various spatial scales through chain-connected ladder-style information flows and fuses them in a two-stage process, namely pre-fusion and re-fusion. The serial flow continuously increases receptive fields of output neurons and those in parallel encode different region-based contexts. Each information flow is a shallow encoder-decoder with appropriate down-sampling scales to sufficiently capture contextual information. We further adopt an attention model in CAM  to guide feature re-fusion. Based on these developments, we construct the Chained Context Aggregation Network (CANet), which employs an asymmetric decoder to recover precise spatial details of prediction maps. We conduct extensive experiments on six challenging datasets, including Pascal VOC 2012, Pascal Context, Cityscapes, CamVid, SUN-RGBD and GATECH. Results evidence that CANet achieves state-of-the-art performance.
\end{abstract}


%
\IEEEpeerreviewmaketitle

\section{Introduction}
\begin{figure}[t]
\begin{center}
\includegraphics[width=1.0\linewidth]{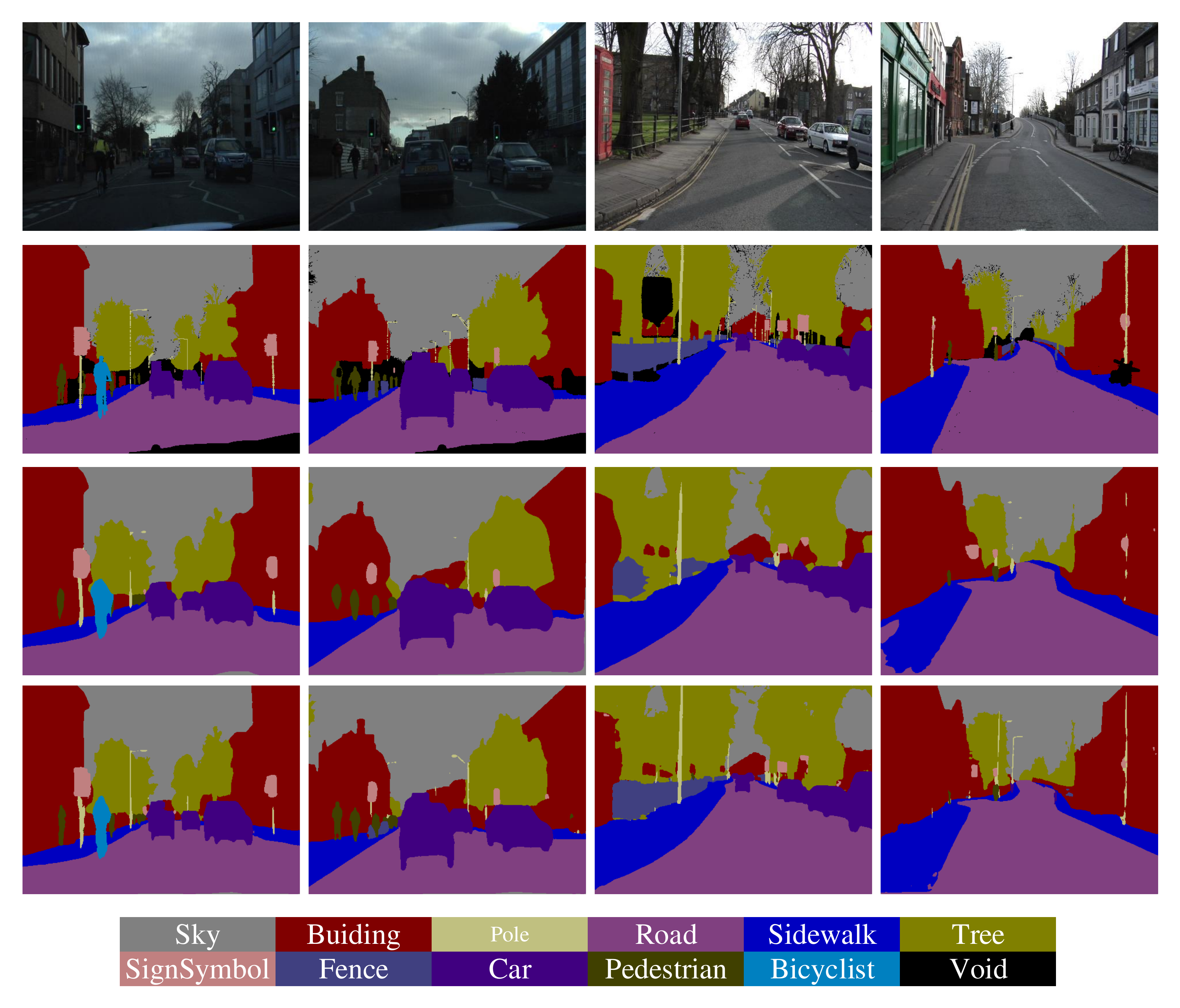}
\end{center}
\caption{Some visualized predictions on CamVid test set. \textbf{First row}: input images. \textbf{Second row}: ground truth. \textbf{Third row}: predictions of dilated FCN. \textbf{Fourth row}: predictions of CANet (ours), which obtains more sharper segmentation boundaries such as poles.}
\label{fig:camvid}
\end{figure}

\IEEEPARstart{S}{emantic} segmentation is a vital task in computer vision, aiming to assign corresponding semantic labels to each pixel in images. It has fundamental applications in the fields of automatic driving~\cite{Cordts_2016_CVPR,pohlen2017full}, medical image~\cite{ronneberger2015u}, augmented reality, etc. Dominant techniques in modern semantic image segmentation are based on the Fully Convolutional Network (FCN)~\cite{long2015fully}, achieving optimal performance in this task. FCN adapts deep image classification models~\cite{he2016deep,simonyan2014very,huang2017densely,chollet2017xception} for dense prediction by replacing fully connected layers with convolutions and gains increasing receptive fields and high-level contexts through cascaded convolutional and sub-sampling pooling layers. However, the continuous down-sampling process causes the loss of spatial details, resulting in poor object delineation and small spurious regions. Fig.~\ref{fig:camvid} shows some examples, where segmented objects present only blurry contours such as the pole and sidewalk. In summary, the paradox between semantics and spatial details is a significant predicament of FCN based approaches.

Combining dilated/atrous convolutions and context modules~\cite{chen2014semantic,yu2015multi,zhao2017pyramid,chen2017rethinking,zhang2018context,he2019adaptive,zhang2019co,fu2019dual,zhu2019asymmetric,zhang2019acfnet} becomes a popular alternative to reconcile the above contradiction. Dilated convolution can increase the receptive field while maintaining feature maps’ resolutions without extra parameters. However, it suffers from the gridding dilemma~\cite{wang2018understanding}, relinquishing part of the neighboring information that is also essential for elaborate semantic segmentation on account of all pixels interacting with surrounding ones to make up objects and form local contexts. Context modules remedy this problem in some way. They join feature maps of various but larger receptive fields to exploit both local and global contexts. Global cues help to understand the entire image scene and, to some extent, reject the ambiguity caused by similar local objects, e.g. cars instead of ships are more likely to show up in a city scene.

Many existing methods~\cite{zhao2017pyramid,chen2017rethinking,he2019adaptive,chen2017deeplab,chen2018encoder,li2018pyramidattention,liu2018semantic} adopt a parallel context module design that encodes contextual information through separate convolutional paths. For example, PSPNet~\cite{zhao2017pyramid} focuses on global contexts and uses large pooling strides, which seriously injures spatial details. Differently, Atrous Spatial Pyramid Pooling (ASPP)~\cite{chen2017rethinking,chen2018encoder} employs dilated filters to exploit sub-region contexts but exacerbates the gridding effect~\cite{wang2018understanding}. Both adopt a single convolutional layer of limited learning ability to transmit features. ACFNet~\cite{he2019adaptive} leverages a global image representation to re-weight sub-region features, which is computationally inefficient. Those methods explore contextual information solely through parallel information flows that lack feature interaction, resulting in a restricted variety of receptive fields and inconsistency of feature scales.

Meanwhile, some networks~\cite{newell2016stacked,fu2017densely,fu2019stacked} employ stacked encoder-decoder structures of sufficient learning ability to exploit contextual information and bring fine localization recovery. We can treat them as in-series context modules where the latter features exclusively depend on the previous. Therefore, such in-series structures lack feature diversity and flexibility intrinsically. Besides, they usually lead to deeper networks that are difficult to train. Empirical receptive field size is not proportional to the depth of networks~\cite{zhou2014object}, on the other hand. To obtain elegant convergence, SDN~\cite{fu2019stacked} introduces inter-unit and intra-unit skip connections and hierarchical supervision, resulting in a sophisticated model construction procedure and training pipeline.

\begin{figure*}
\begin{center}
\includegraphics[width=1.0\linewidth]{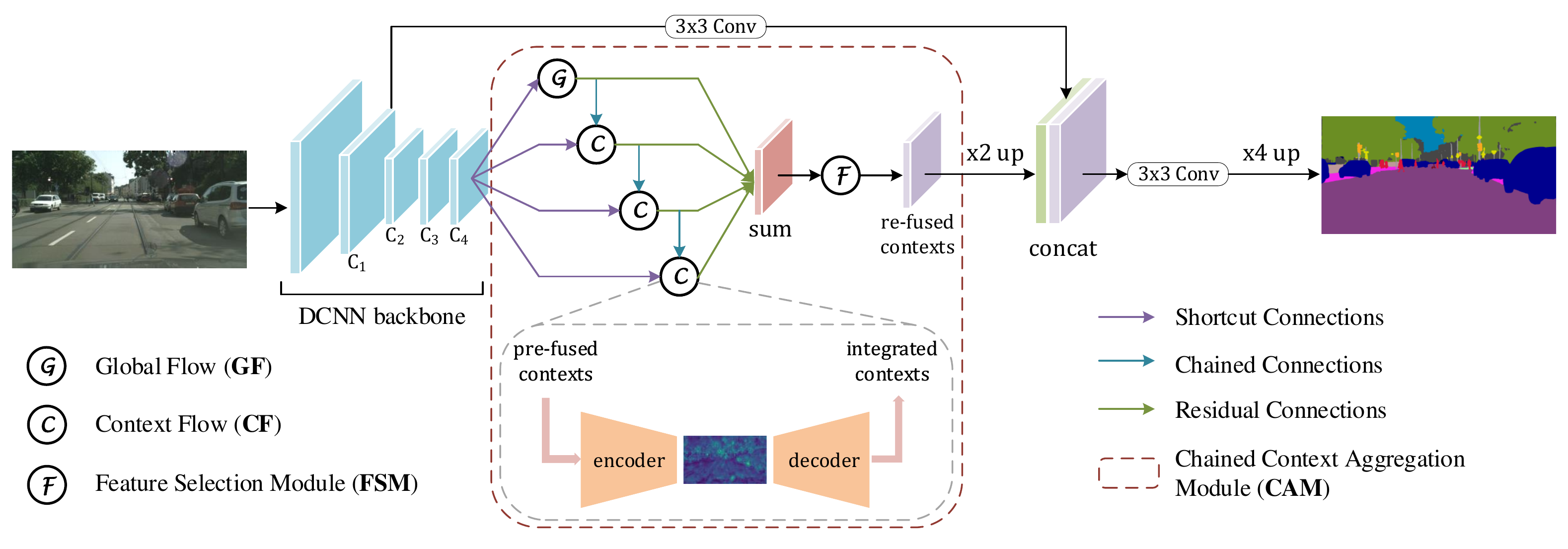}
\end{center}
\caption{Overview of the proposed CANet. Given an input image, we first adopt a deep convolutional neural network (DCNN) as the backbone to encode a shared feature map, and then the carefully designed Chained Context Aggregation Module (CAM) is applied to enrich multi-scale contexts, followed by an asymmetric decoder to get the final per-pixel prediction. ``$\times N$ up'' means $N$-time up-sampling operation. Shortcut connections, Chained Connections and Residual Connections are collectively called Flow Guidance Connections.}
\label{fig:overview}
\end{figure*}

To address the above obstacles,  we propose an innovative paradigm termed the Chained Context Aggregation Module (CAM) with effectiveness and clearness, as illustrated in Fig.~\ref{fig:overview}. CAM utilizes Flow Guidance Connections to connect multiple information flows in a series-parallel hybrid manner. Each flow is a shallow encoder-decoder with suitable down-sampling scales to integrate contextual information. Global Flow (GF) aims to obtain a global receptive field, and Context Flow (CF) captures sub-region based contexts of individual scales. The serial GF and CFs (i.e. several Context Flows) guided by Chained Connections contain multiple encoder-decoder blocks, increasing the receptive fields of output neurons to construct contextual information and recover localization information. Parallel GF and CFs encode contexts of different spatial scales to obtain potentially accurate feature maps for multi-scale objects segmentation. 

Within the series-parallel hybrid architecture, a two-stage feature fusion mechanism is naturally developed to strengthen context interaction, namely pre-fusion and re-fusion. As an extension work, we sharpen the simple yet effective decoder of DeepLabv3+~\cite{chen2018encoder} with a compact $3\times3$ convolution to intensify semantic representations of low-level features. Based on these developments, we construct the Chained Context Aggregation Network (CANet) for semantic image segmentation and conduct extensive experiments on six challenging datasets whose results demonstrate the superiority.

We conclude the critical contributions as follows:
\begin{itemize}
    \item We propose the CAM to capture multi-scale contexts via multiple information flows in a series-parallel hybrid manner. Each information flow is a shallow encoder-decoder to integrate contextual information with sufficient learning ability.
    \item The serial flow increasingly enlarges receptive fields of output neurons while parallel flows encode different region-based contexts. Both are of significant importance to obtain potentially accurate representations and improve model performance considerably.
    \item Flow Guidance Connections enable sufficient feature interaction of multiple information flows and powerful aggregation of multi-scale contexts through a naturally developed pre-fusion/re-fusion process. They are crucial for model convergence with succinctness at the same time. We further utilize attention models in CAM to boost feature re-fusion and refine the results.
    \item We construct a generalized framework termed the CANet and achieve state-of-the-art performance on the benchmarks of Pascal VOC 2012, Pascal Context, Cityscapes, CamVid, SUN-RGBD and GATECH.
\end{itemize}

The remainder of this paper is organized as follows. We briefly review related works in Section~\ref{section:related_work}. In Section~\ref{section:method}, we detail the proposed CANet and represent high-level functions behind the critical series-parallel hybrid CAM. Section~\ref{section:experiments} presents quantitative experimental results that verify the superiority of the proposed framework. Finally, we summarize our work in Section~\ref{section:conclusion}.

\section{Related Work}
\label{section:related_work}
With increasing applications of deep learning methods to semantic segmentation in recent years, the task has made breakthroughs on benchmarks. We briefly review related research works in this section.

\subsection{Spatial Information}
In convolutional networks, each layer handles diverse information. Low-level layers usually have more positional information and high-level ones hold more semantics. Both positions and semantics play a pivotal role in semantic segmentation. FCN~\cite{long2015fully} based methods obtain high-level semantics through down-sampling operations at the expense of spatial details, leading to the problems of poor boundary positioning and inconsistent predictions. F. Yu and V. Koltun~\cite{yu2015multi} develop dilated convolutions to reduce spatial loss during encoding and achieve excellent results. Following this idea, PSPNet~\cite{zhao2017pyramid}, DeepLabv3+~\cite{chen2018encoder}, APCNet~\cite{he2019adaptive}, DANet~\cite{fu2019dual} and CFNet~\cite{zhang2019co} all apply dilated convolutions in the backbone network to maintain the resolution of feature maps. FRRN~\cite{pohlen2017full} and HRNet~\cite{sun2019deep} employ a high-resolution branch to serve the purpose.

Another idea utilizes the encoder-decoder structure to restore resolution via up-sampling or deconvolution. Typically, they contain an encoder that down-samples the input image to obtain high-level semantics and a decoder that gradually restores the resolution to classify every pixel. Both SegNet~\cite{badrinarayanan2017segnet} and UNet~\cite{ronneberger2015u} utilize a symmetric decoder to obtain fine-recovered predictions. GCN~\cite{peng2017large} and RefineNet~\cite{lin2017refinenet} are further developments with carefully designed decoders. SDN~\cite{fu2019stacked} exploits contextual information by stacking multiple encoder-decoders. However, these methods significantly increase the model's depth, introducing a huge parameter size and aggravating the gradient degradation problem. Deconvolution~\cite{noh2015learning} and DUpsample~\cite{tian2019decoders} employ up-sampling strategies different from bilinear interpolation as decoders to achieve better results.

\subsection{Multi-scale Contexts}
Existing context-based methods aim to capture effective context features that play a vital role in segmenting multi-scale objects. At the top of the backbone network, PSPNet~\cite{zhao2017pyramid} and DeepLab~\cite{chen2017rethinking,chen2017deeplab,chen2018encoder} employ parallel information flows to perceive multi-scale representations by different pooling strides and dilation rates, respectively. APCNet~\cite{he2019adaptive},  DANet~\cite{fu2019dual} and CFNet~\cite{zhang2019co} take advantage of attention models to obtain various contextual information. Based on the U-shape structure, RefineNet~\cite{lin2017refinenet}, LFNet~\cite{yu2018learning} and GCN~\cite{peng2017large} achieve a productive fusion of hierarchical features.  DFANet~\cite{li2019dfanet} implements an in-depth aggregation of hierarchical features by cascading multiple encoders. Besides, Recurrent Neural Networks (RNNs) are also developed to capture long-range dependencies~\cite{shuai2016dag,fan2018multi}. In this work, we evaluate semantic image segmentation from a novel spatial perspective.

\subsection{Attention Mechanism}
The core idea of attention mechanism is to assign distinctive attention weights to different parts of the input, just like people focusing on attractive parts of seeing features. Like the non-local block~\cite{wang2018non} introducing self-attention~\cite{vaswani2017attention} into computer vision, various types of attention mechanisms~\cite{hu2018squeeze,Li_2019_CVPR,woo2018cbam,fu2017look} play an increasingly important role in this field. Methods with attention models for semantic segmentation~\cite{zhang2018context,he2019adaptive,zhao2018psanet,pang2019towards,zhang2019acfnet,fu2019dual,huang2019ccnet,zhu2019asymmetric} also further improve the performance. EncNet~\cite{zhang2018context} and ACFNet~\cite{he2019adaptive} leverages a global image representation as guidance to estimate the local affinity coefficients. ACFNet~\cite{zhang2019acfnet} extracts global contexts via computing class centers. DANet~\cite{fu2019dual} and CCNet~\cite{huang2019ccnet} are both successful instantiations of the non-local network~\cite{wang2018non} in semantic segmentation. Those attention-based methods bring various feature fusion mechanisms via adaptively learned attention weights. Differently, this paper mainly focuses on the feature encoding paradigm through diversifying feature propagation and interaction to encode multi-scale contexts.

\section{Method}
\label{section:method}
We propose the Chained Context Aggregation Network to enable flexible capturing and aggregation of multi-scale contextual information and explore its improvement for semantic segmentation. We elaborate on the framework in this section. 

\subsection{Overview}
As shown in Fig.~\ref{fig:overview}, we design the Chained Context Aggregation Module (CAM) to diversify feature propagation and encode multi-scale contextual information over feature embedding extracted by a deep convolutional neural network (DCNN). To be specific, following prior works~\cite{zhao2017pyramid,zhang2018context,chen2017rethinking}, we employ ResNet~\cite{he2016deep} with the dilation strategy~\cite{chen2014semantic} as the DCNN backbone to preserve spatial resolution without extra parameters. At the top of the backbone lies the carefully designed CAM, where several shallow encoder-decoders serve as information flows to integrate contextual information at individual scales. CAM utilizes Flow Guidance Connections to develop a series-parallel hybrid structure of information flows, expecting to exploit multi-scale contextual features effectively to improve segmentation performance for objects of different spatial sizes. We further adopt a channel-attention model called the Feature Selection Module (FSM) to promote feature fusion. Finally, CANet mimics the simple yet effective asymmetric decoder proposed by DeepLabv3+~\cite{chen2018encoder} to up-sample the prediction maps to present classification confidence for each class at every pixel. We note that CANet replaces the original $1\times1$ convolution with $3\times3$ filters in the decoder to reduce the semantic gap between low- and high-level pixel features. CANet utilizes bilinear interpolation as a naive up-sampling strategy.


\begin{figure}[t]
\begin{center}
\includegraphics[width=1.0\linewidth]{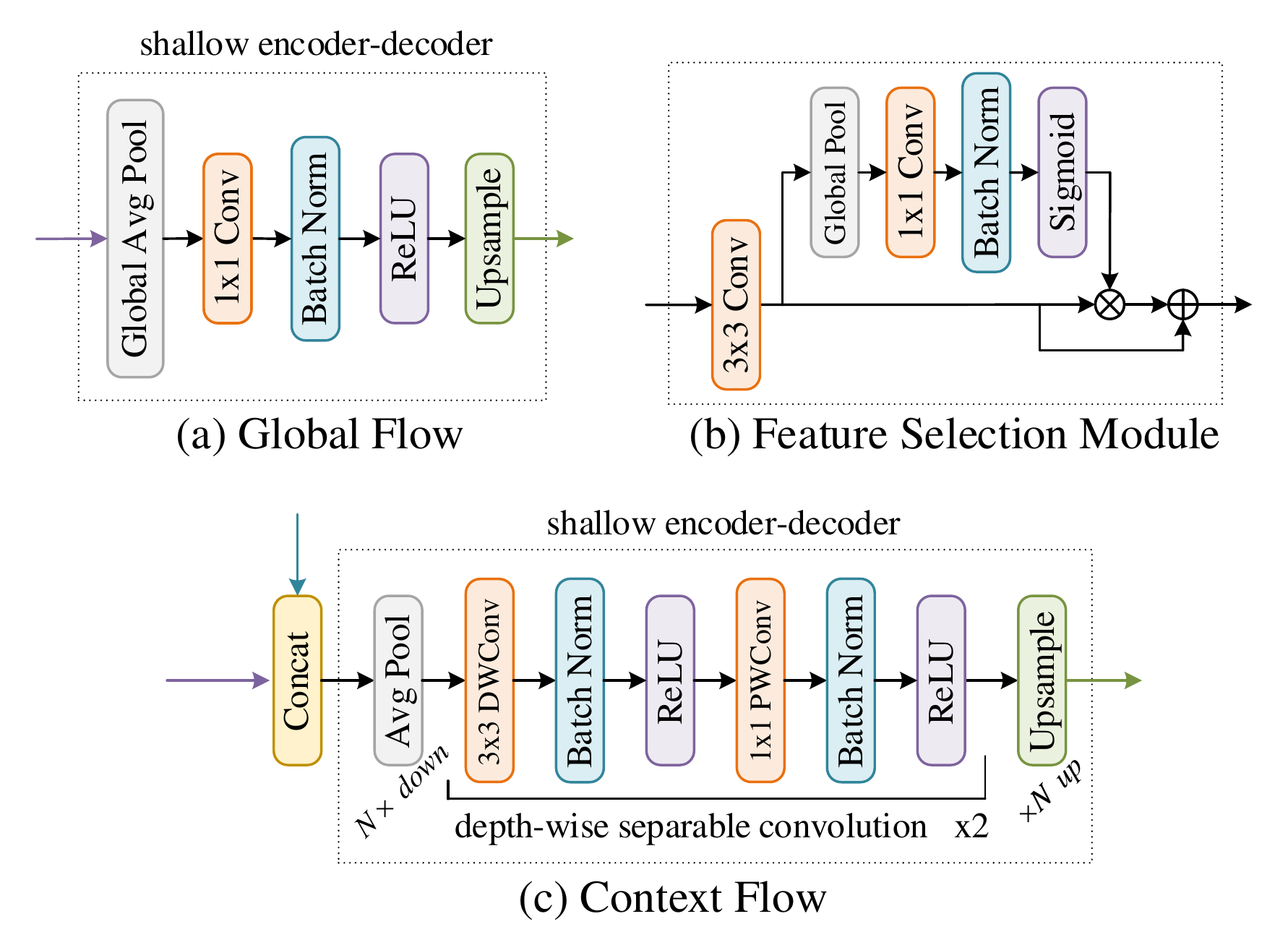}
\end{center}
\caption{Detailed components of CAM. ``$N \times$ down'' means $N$-time down-sampling of the inputs by average pooling layer, while ``$\times N$ up'' means $N$-time up-sampling operation. DWConv and PWConv are shorthand for depth-wise convolution and point-wise convolution respectively. We stack two depth-wise separable convolutions to integrate pre-fused features and enlarge receptive fields. $\oplus$ and $\otimes$ represents element-wise summation and element-wise product, respectively.}
\label{fig:cam_details}
\end{figure}

\subsection{Chained Context Aggregation Module}
The CAM is the critical part of CANet to aggregate multi-scale contextual information. Based on the backbone network's shared features, CAM further exploits semantic relations at different spatial scales through GF and CF. Both are shallow encoder-decoder structures containing a down-sampling layer to gain different receptive field, a projection layer to integrate sufficient context features, and an up-sampling layer to recover localization information. We unite GF and CFs in a series-parallel hybrid structure to diversify feature propagation and encode multi-scale contexts.

The series-parallel hybrid design brings four advantages. Firstly, shallow encoder-decoders of sufficient learning ability serve as information flows to integrate contextual information, yielding fine spatial resolution recovery than a single convolutional layer. Secondly, serial GF and CFs continuously enlarges receptive fields of output neurons and enhances semantic representations. At the same time, it deepens the network, thus increases learning ability. Thirdly, parallel flows encode pixel features at different scales, propitious for multi-scale object segmentation. Finally, such design diversifies feature propagation and facilitates the back-propagation of gradients with simplicity, making the model easy to train.

Within the series-parallel hybrid architecture, a two-stage feature fusion mechanism is naturally developed. It diversifies feature transmission and alleviates scale inconsistency among different flows. More specifically, CF first concatenates shared features of the backbone and output features of the upper flow and integrate them through a shallow encoder-decoder. We name the process pre-fusion. Residual Connections guide the aggregation of different information flows, followed by an attention model to produce rich contextual information, which we name as re-fusion.

\begin{figure}[t]
\begin{center}
\includegraphics[width=1.0\linewidth]{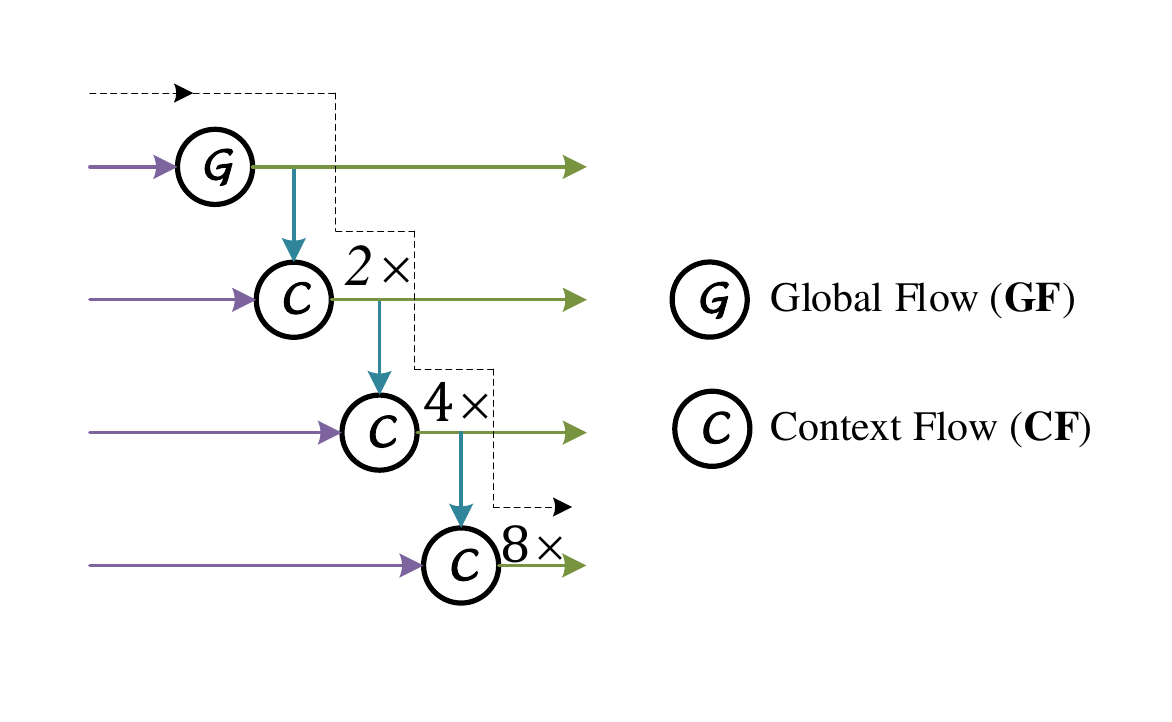}
\end{center}
\caption{One possible combination of GF and CFs. ``$N \times$'' indicates the down-sampling scales of CFs. The black dotted arrow points out the direction of multiple in-series encoder-decoder flows.}
\label{fig:cfs_example}
\end{figure}

We can see that during the above process, contexts obtained by a latter information flow do not entirely depend on that of the previous. It is the series-parallel structure that enables robust multi-scale context aggregation. Different combinations of different quantity and down-sampling scales of CFs make up diverse CAM to exploit contexts of various scales. It is another embodiment of CAM's flexibility. Fig.~\ref{fig:cfs_example} provides a possible combination where the information flow symbolized by the black dotted arrow makes up a stack of multiple shallow encoder-decoders that can be regarded as a particular case of SDN~\cite{fu2019stacked}. The main components of CAM are described below.

\subsubsection{Global Flow}
Practices~\cite{zhao2017pyramid,chen2017rethinking,zhang2019co,liu2015parsenet} have witnessed that global pooling features can provide a global receptive field as a reliable cue to distinguish confusing objects. In CAM, it is achieved by applying global average pooling on shared feature maps of the backbone network, which we refer to as Global Flow. Fig.~\ref{fig:cam_details}(a) depicts its details where we employ a single $1\times1$ convolution to obtain global image representations.

\subsubsection{Context Flow}
We propose Context Flow to exploit local contexts of different receptive fields, as shown in Fig.~\ref{fig:cam_details}(c). CF is the spotlight where pre-fusion occurs. More specifically, given the two inputs of different spatial scales coming from the shared backbone and upper flow, CF first concats them in channel dimension and down-samples $N\times$ through an average pooling layer to capture sub-region based contexts, which also lessens computation cost and reduces information redundancy. The following two consecutive convolutional layers serve as the projection layer, eliminating the aliasing effect and encode integrated context features, where depth-wise separable convolutions~\cite{howard2017mobilenets} are adopted to diminish model parameters. Finally, CF up-samples the output to the same size of the input. By setting a suitable down-sampling scale $N$, we can obtain different receptive fields and accurate position information through the encoder-decoder paradigm.

\subsubsection{Flow Guidance Connections}
GF and CFs are devised to obtain global and local contexts of various spatial scales, respectively. We propose Flow Guidance Connections, the principal notion of forming the series-parallel hybrid structure, to unite GF and CFs to enhance feature delivery and enrich multi-scale contexts. Flow Guidance Connections include Shortcut Connections, Chained Connections and Residual Connections, as separately depicted by the purple, blue and green arrows in Fig.~\ref{fig:overview}. Shortcut Connections let CAM reuse the pixel embeddings of the backbone network. They not only decouple lower and upper features but also efficiently promote the acquisition and diversity of multi-scale contexts. Since GF and CFs have distinct receptive fields, Chained Connections intend to guide the pre-fusion that magnifies the flexible feature aggregation and decreases feature inconsistencies between adjacent information flows. Finally, Residual Connections serve as ushers of the re-fusion process and also alleviate gradient vanishing caused by the serial flow increasing the network’s depth, which makes the whole framework easy to train. Aggregated feature maps are then fed into the FSM to construct re-fused contexts.

\subsubsection{Feature Selection Module}
Although feature pre-fusion alleviates the scale inconsistency to a certain extent, we still adopt an attention model to guide feature re-fusion. Inspired by SENet~\cite{hu2018squeeze}, we adopt a simple channel-attention based FSM to select advantageous features and suppress the useless or harmful at the channel level, balancing effects of different region-based contexts on segmentation results. Assume the integrated local contexts of the $i$-th CF is $\bm{X}_i$ and $\bm{G}$ denotes global contexts obtained by GF, then
\begin{equation}
    \bm{U}=F_{\mathit{bilinear}}(\bm{G})+\sum_{i}F_{\mathit{bilinear}}(\bm{X}_i)
\end{equation}
\begin{equation}
    \bm{U}'=\delta(F_{\mathit{BN}}(f(\bm{U},\bm{W}_1)))
\end{equation}
\begin{equation}
    \widetilde{\bm{U}}=\bm{U}' \otimes \sigma(F_{\mathit{BN}}(f(F_{\mathit{GAP}}(\bm{U}'),\bm{W}_2))) \oplus \bm{U}'
\end{equation}
where $F_{\mathit{bilinear}}$ is bilinear interpolation function, $F_{\mathit{BN}}$ and $F_{\mathit{GAP}}$ the normalization layer and global average pooling layer, respectively. $f(\cdot,\cdot)$ denotes convolution. $\sigma$ and $\delta$ refer to the Sigmoid and ReLU function. $\bm{W}_1$ and $\bm{W}_2$ are both learnable parameters. $\otimes$ represents element-wise product and $\oplus$ element-wise summation. $\widetilde{\bm{U}}$ is the re-fused contexts, which are then up-sampled by the decoder to obtain predictions. Fig.~\ref{fig:cam_details}(b) illustrates the details.

\subsection{Loss Function}
\label{subsection:loss}
We employ standard cross-entropy loss for training, as shown in Eq.~\ref{eq:ce_loss}. Since CAM further deepens the network, here we use an auxiliary loss to serve for a better backbone convergence and a principal loss to supervise the output of the whole network, as Eq.~\ref{eq:joint_loss} and~\ref{eq:joint_loss_2} present.
\begin{equation}
    H(\bm{y},\bm{\hat{y}})=-\frac{1}{N}\sum_{i}y_i\log{\hat{y}_i}
\label{eq:ce_loss}
\end{equation}
where $\bm{y}$ denotes the ground-truth and $\bm{\hat{y}}$ the prediction of network.
\begin{equation}
    \bm{y}_c=F_\mathit{class}(\bm{c}_3)
\label{eq:joint_loss}
\end{equation}
\begin{equation}
    \mathcal{L}=H(\bm{y},\bm{y}_d)+\lambda H(\bm{y},\bm{y}_c)
\label{eq:joint_loss_2}
\end{equation}

where $\bm{y}_d$ is the output of the decoder, $\bm{c}_3$ the intermediate output of the backbone as shown in Fig.~\ref{fig:overview}, and $F_\mathit{class}$ a classification layer with $3\times3$ convolution. $\lambda$ is adopted to balance the training process. We do not use the auxiliary output when inference.

\section{Experiments and Results}
\label{section:experiments}
We conduct extensive experiments on Pascal VOC 2012~\cite{everingham2010pascal}, Pascal Context~\cite{mottaghi2014role}, Cityscapes~\cite{Cordts_2016_CVPR}, CamVid~\cite{brostow2009semantic}, SUN-RGBD~\cite{song2015sun} and GATECH~\cite{hussain2013geometric} to evaluate the performance of our proposed CANet. Results are obtained with multi-scale and flipping skills if not specified. We adopt the standard benchmarks, pixel accuracy (PA) and mean intersection over union (mIoU) as the evaluation metrics. We use only mIoU on Pascal VOC 2012, Pascal Context and Cityscapes for common convention. We assume segmentation label space $\mathcal{S}=\{l_0,l_1,...,l_k\}$ with a total of $k+1$ classes where $l_0$ is the background or a void class. $p_{ij}$ represents the amount of pixels of class $i$ but inferred to be class $j$, then
\begin{equation}
    \mathit{PA}=\frac{\sum_{i=0}^kp_{ii}}{\sum_{i=0}^k\sum_{j=0}^kp_{ij}}
\label{eq:pa}
\end{equation}
\begin{equation}
    \mathit{mIoU}=\frac{1}{k+1}\sum_{i=0}^k{\frac{p_{ii}}{\sum_{j=0}^kp_{ij}+\sum_{j=0}^kp_{ji}-p_{ii}}}
\label{eq:miou}
\end{equation}

We implement experiments by MXNet~\cite{chen2015mxnet} and borrow ImageNet~\cite{russakovsky2015imagenet} pre-trained backbone from the open-source toolkit GluonCV~\cite{guo2019gluoncv}. We set the dilation rates of backbone’s last two residual modules to 2 and 4, respectively. Thus the resolution of backbone’s final output feature map is 1/8 of the input image. Following T. He et al.~\cite{he2019bag}, we replace the original $7\times7$ convolution with three stacked $3\times3$ convolutional layers in ResNet. The output channels of FSM are 256. Like DeepLabv3+~\cite{chen2018encoder}, we set output channels of the $3\times3$ convolution in the decoder to 48 to prevent low-level features from outweighing the importance of the rich contextual information. We adopt ResNet101~\cite{he2016deep} as the backbone for comparisons with prior methods.

We use the poly learning rate policy $\mathit{lr}=\mathit{baselr}\times (1-\frac{\mathit{iter}}{\mathit{total\_iter}})^{\mathit{power}}$ and set the power to 0.9. The initial learning rate is 4e-3 for Cityscapes and 1e-3 for the other five datasets. We use the standard mini-batch stochastic gradient descent (SGD) as the optimizer and set momentum to 0.9. To prevent over-fitting, we set the weight decay to 5e-4 for CamVid and GATECH, and 1e-4 for the others. For data augmentation, we first flip input images with a probability of 0.5 and randomly scale them from 0.5 to 2.0 times. Then we crop the images with padding if needed. Finally, a random Gaussian blur is added. Since appropriate crop size influences the model performance, we  empirically crop images to $384\times 384$ for CamVid, $768\times768$ for Cityscapes, and $480\times480$ for the others.

\subsection{Results on Pascal VOC 2012}

\begin{table}
\caption{Investigation of the quantity and down-sampling scales of CFs}
\begin{center}
\begin{tabular}{lc|c}
\toprule
\textbf{Backbone} & \textbf{CFs} & \textbf{mIoU(\%)} \\
\midrule\midrule
ResNet50 & None(baseline) & 69.97 \\
ResNet50 & \{2\} & 77.80 \\
ResNet50 & \{2, 2\} & 78.10 \\
ResNet50 & \{2, 4\} & 78.07 \\
ResNet50 & \{2, 2, 2\} & 78.29 \\
ResNet50 & \{2, 4, 4\} & 78.33 \\
ResNet50 & \{2, 4, 8\} & 78.28 \\
ResNet50 & \{2, 2, 2, 2\} & 78.51 \\
ResNet50 & \{2, 4, 8, 16\} & \textbf{78.68} \\
ResNet50 & \{2, 4, 8, 16, 32\} & 77.88 \\
\bottomrule
\end{tabular}
\end{center}
\label{table:ablation_cfs}
\end{table}

\begin{figure}[t]
    \centering
    \subfigure[]{
        \begin{minipage}[t]{0.23\textwidth}
        \includegraphics[width=\textwidth]{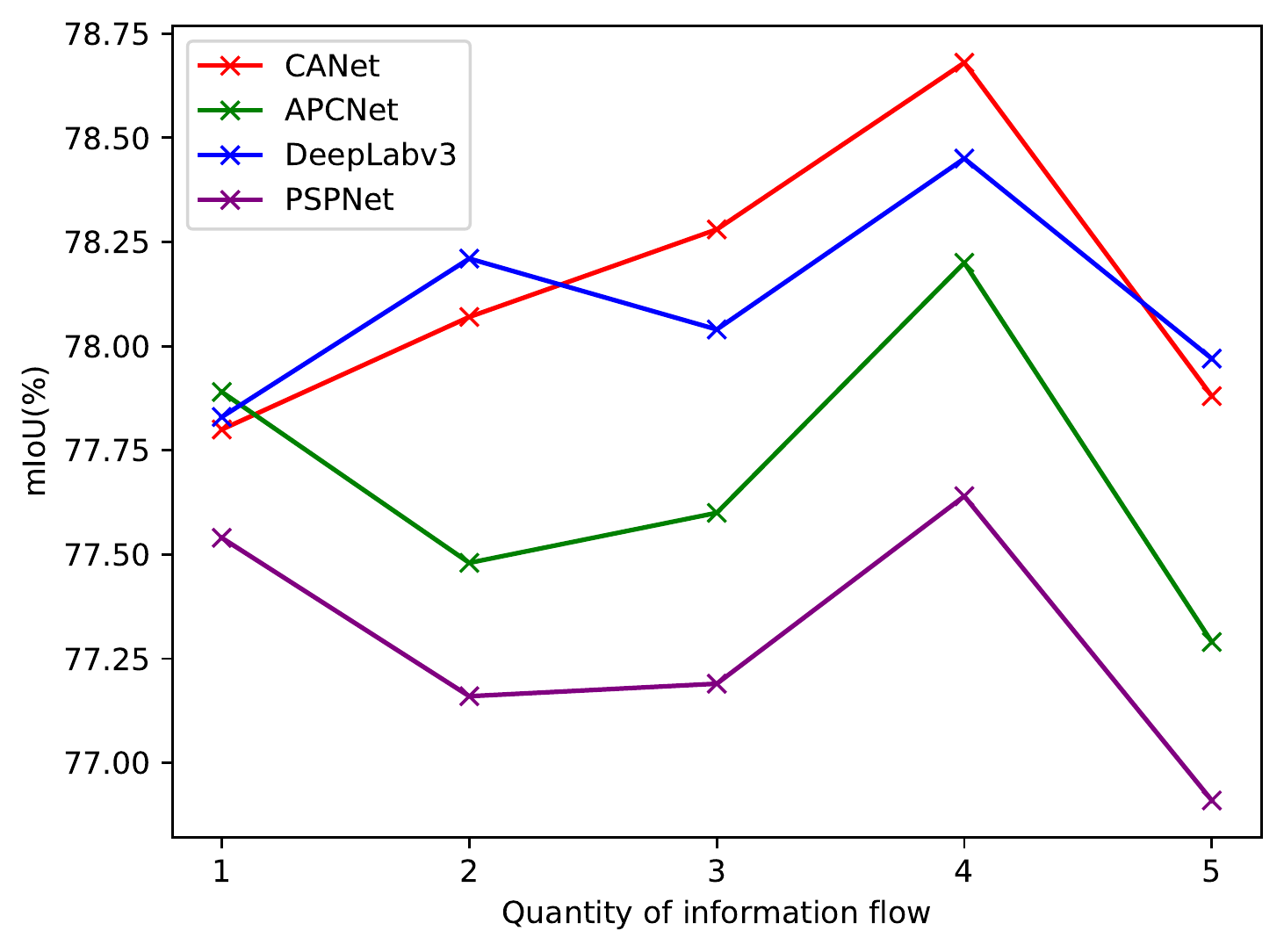}
        \end{minipage}}
    \subfigure[]{
        \begin{minipage}[t]{0.23\textwidth}
        \includegraphics[width=\textwidth]{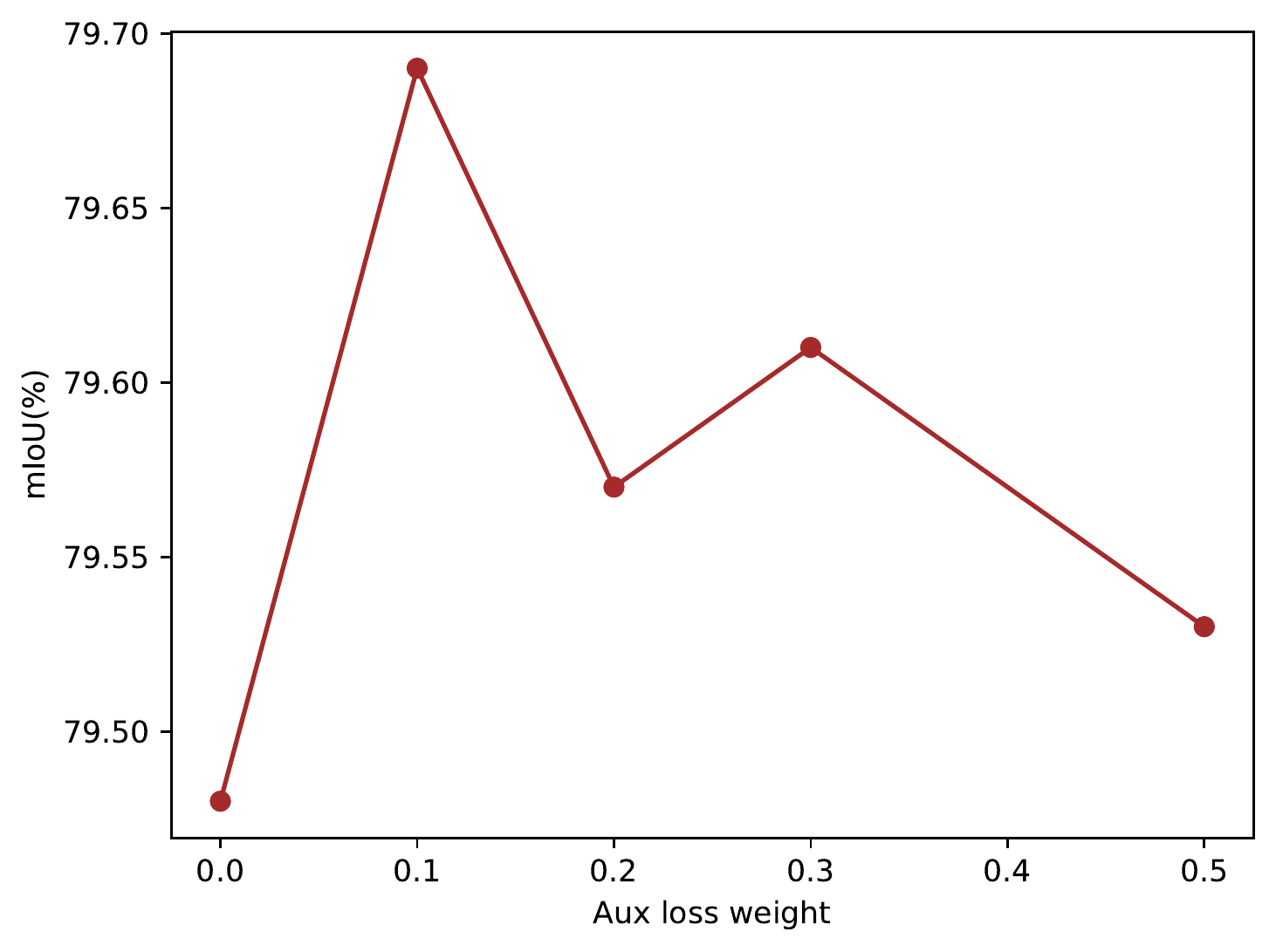}
        \end{minipage}}
        \caption{(a) The number of information flow and mIoU performance on Pascal VOC 2012 validation set. Since CFs are proposed to capture sub-region based contexts, we do not take into account global average pooling paths of CANet, DeepLabv3 and PSPNet when counting the number. The $1\times1$ convolution projection in ASPP of DeepLabv3 is also ignored. Down-sampling scales \{2, 4, 8, 16, 32\}, dilation rates \{6, 12, 18, 24, 36\}, pyramid scales \{1, 2, 3, 6, 32\} and output bin sizes of pooling operations \{2, 3, 6, 32, 48\} are set for CANet, DeepLabv3, APCNet and PSPNet, respectively. (b) Ablation study of auxiliary loss weight $\lambda$ as stated in Section~\ref{subsection:loss}.}
        \label{fig:comparisons}
\end{figure}

Pascal VOC 2012~\cite{everingham2010pascal} contains 1464, 1449 and 1456 images for training, validation and testing respectively. All images are pixel-wise labeled with 21 semantic classes, one of which is background. Following prior works~\cite{long2015fully,zhao2017pyramid,chen2014semantic}, we augment the training set with SBD dataset~\cite{hariharan2015hypercolumns} for experiments, resulting in 10582 images for training. We first perform sound ablation studies on the validation set to verify the benefits of key ideas in CANet as well as to explore the improvement of different combinations of CFs on the segmentation results. Our baseline is dilated ResNet based FCN~\cite{long2015fully,he2016deep,chen2017deeplab}. All ablation results are based on single scale inputs.

\subsubsection{Down-sampling Scales of CFs}
We believe that different quantities and down-sampling scales of CFs contribute to capturing different scales of contextual information of objects. Fig.~\ref{fig:cfs_example} gives a possible example of CFs. We conduct some exploratory experiments on this and Table~\ref{table:ablation_cfs} reports the results, where $\{d_1,d_2,…,d_n\}$ means there are $n$ CFs chain-connected from top down, and the down-sampling scales are  $d_1,d_2,...,d_n$ respectively. It can be seen that:
\begin{itemize}
    \item Compared to the baseline FCN, the segmentation performance is considerably improved no matter what combination of CFs is. And the \{2, 4, 8, 16\} achieves the best with an 8.71\% mIoU improvement. This observation suggests that CAM is contributory to accurate segmentation of multi-scale targets. Besides, different combinations introduce non-negligible performance perturbation. 
    \item CAM gains the best at the quantity of 4 CFs, but the performance degrades noticeably when the number of CFs goes beyond. To this end, we further investigate three representative state-of-the-art methods of PSPNet~\cite{zhao2017pyramid}, DeepLabv3~\cite{chen2017rethinking} and APCNet~\cite{he2019adaptive} that adopt similar multiple information flow designs. The relationship between the quantity and model performance is shown in Fig.~\ref{fig:comparisons}(a). Although those methods hold different interpretations of information flows, the results show compatibility. We consider that the increase of flows brings more diverse region-based contexts and improves model performance. However, scale inconsistency increases at the same time, which dilutes semantics during feature fusion and causes performance degradation after exceeding a specific limit.
    \item Under the same quantity of CFs, more flexible down-sampling rates combination yields more diverse local contexts, leading to a slightly better performance.
\end{itemize}

In all the following experiments, we adopt the down-sampling scales of \{2, 4, 8, 16\}.

\begin{figure*}
\begin{center}
\includegraphics[width=1.0\linewidth]{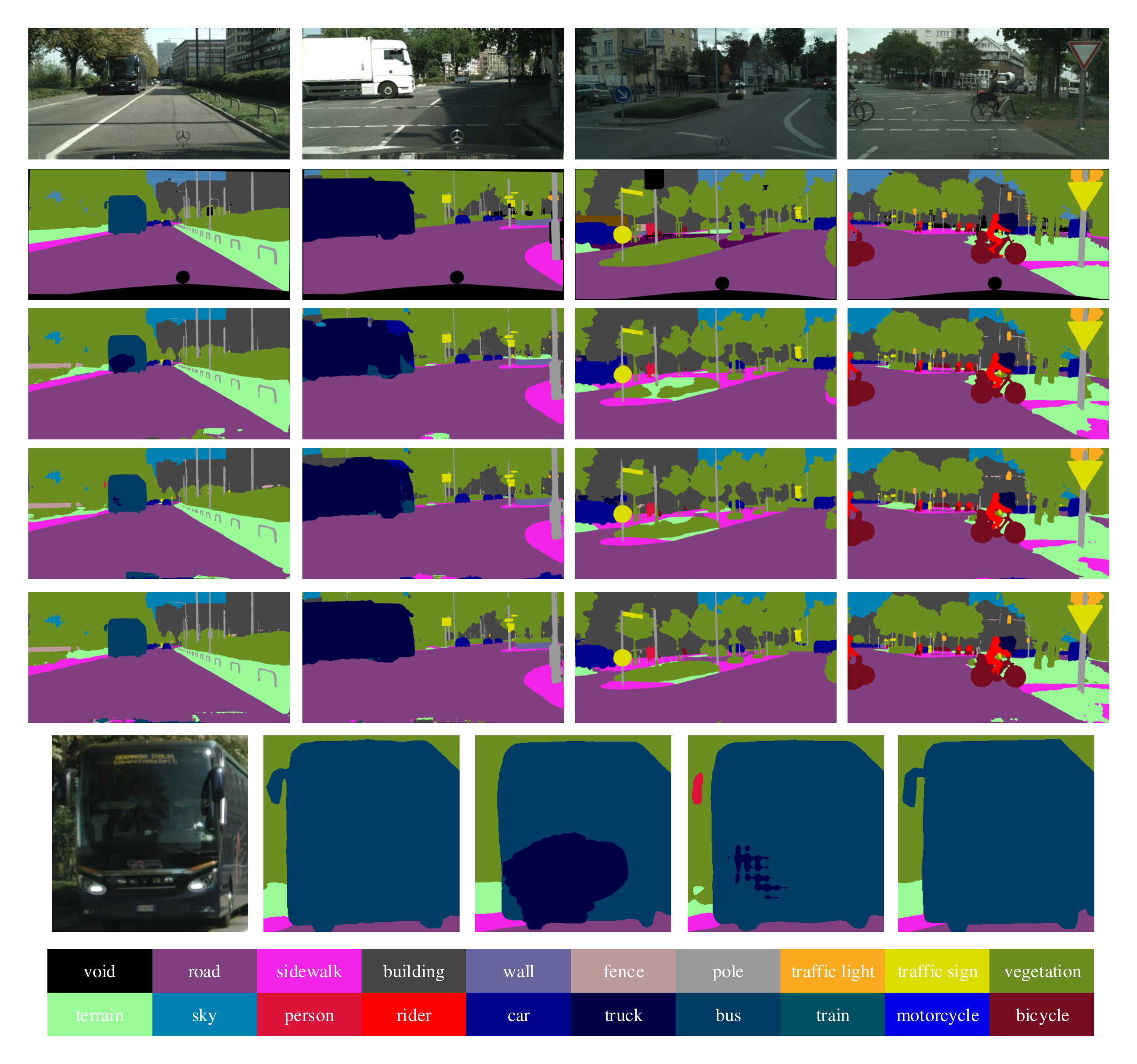}
\end{center}
\caption{Visualized comparisons on the Cityscapes val set. \textbf{First row}: input images. \textbf{Second row}: ground truth. \textbf{Third row}: predictions of PSPNet. \textbf{Fourth row}: predictions of DeepLabv3. \textbf{Fifth row}: predictions of the proposed CANet. We zoom in the first-column example at the last row for better illustration. CANet employs shallow encoder-decoder information flows to integrate context features, which provides fine recovery of localization information and eliminates the gridding effect.}
\label{fig:citysval}
\end{figure*}

\begin{table}
\caption{Investigation of the width of information flows}
\begin{center}
\begin{tabular}{lc|c}
\toprule
\textbf{Backbone} & \textbf{width} & \textbf{mIoU(\%)} \\
\midrule\midrule
ResNet50 & 256 & 78.13 \\
ResNet50 & 384 & 78.36 \\
ResNet50 & 512 & \textbf{78.68} \\
ResNet50 & 768 & 78.64 \\
ResNet50 & 1024 & 78.53 \\
\bottomrule
\end{tabular}
\end{center}
\label{table:width_flow}
\end{table}

\begin{table}
\caption{Validation of the critical series-parallel hybrid design based on dilated ResNet50}
\begin{center}
\resizebox{0.9\linewidth}{!}{
\begin{tabular}{llc|c}
\toprule
\textbf{Method} & \textbf{Design Manner} & \textbf{\# params} & \textbf{mIoU(\%)} \\
\midrule\midrule
    PSPNet~\cite{zhao2017pyramid} & in parallel & 65.7M & 77.64 \\
    DeepLabv3~\cite{chen2017rethinking} & in parallel & 39.8M & 78.45 \\
    APCNet~\cite{he2019adaptive} & in parallel & 69.9M & 78.20 \\
    DenseASPP~\cite{yang2018denseaspp} & densely connected & 50.3M & 78.03 \\
\midrule
    CANet-S & in series & 28.7M & 53.39 \\
    CANet-P & in parallel & 31.9M & 77.90 \\
    CANet & series-parallel hybrid & \textbf{33.0M} & \textbf{78.68} \\
\bottomrule
\end{tabular}}
\end{center}
\label{table:flow_guidance}
\end{table}

\subsubsection{Width of Information Flow}
The width of information flows determines the learning ability of CAM, which is an indispensable parameter for encoding multi-scale features. We verify the impact of this hyper-parameter on model performance and report results in Table~\ref{table:width_flow}. The results indicate that model performance saturates at 512. Larger width may result in over-fitting. We pick this value in all other experiments.

\subsubsection{Validation for the Series-parallel Hybrid Design}
Flow Guidance Connections is a fundamental design of CAM that connect multiple information flows with the critical series-parallel hybrid manner and facilitate model training. To verify the advanced nature of the design, we perform ablation on these connections. We only retain the uppermost Shortcut Connection and the lowermost Residual Connection to form a purely in-series model that is denoted as CANet-S. Besides, we remove all Chained Connections to build CANet-P with a purely parallel structure. We report quantitative comparisons to four similar representative methods in Table~\ref{table:flow_guidance}. Our CANet surpasses all the other context modules with fewer model parameters, which evidences the superiority of the series-parallel hybrid design. It is the design instead of incremental parameters that obtain performance advancement.

As illustrated in Fig.~\ref{fig:cfs_example}, CANet-S only retains the feature propagation flow represented by the black dotted line, which can be seen as an SDN~\cite{fu2019stacked} variant. The difference is that we use no complex cross-layer connections and hierarchical supervision during training. As a result, CANet-S ends up with terrible performance of 53.39\% mIoU. See Table~\ref{table:flow_guidance}. This confirms the contributory to good convergence of the simple yet effective Shortcut Connections and Residual Connections on the other hand. Furthermore, CANet (with pre-fusion and re-fusion) outperforms CANet-P (with only re-fusion) by 0.78 points mIoU, suggesting that the two-stage feature fusion mechanism alleviates feature inconsistency of adjacent information flows, which enhance feature aggregation.

\begin{table*}[]
\caption{Comparison in terms of per-class scores with semblable state-of-the-art baselines on the Cityscapes val set}
\begin{center}
    \resizebox{1.0\linewidth}{!}{
    \begin{tabular}{l|ccccccccccccccccccc|c}
    \toprule
        Method & road & s.walk & build. & wall & fence & pole & t-light & t-sign & veg & terrain & sky & person & rider & car & truck & bus & train & motor & bike & mIoU(\%) \\
    \midrule\midrule
        Dilated FCN~\cite{long2015fully} & 97.8 & 82.7 & 91.8 & 52.0 & 59.1 & 60.9 & 70.3 & 78.2 & 92.0 & 60.9 & 94.1 & 81.1 & 59.8 & 94.3 & 61.7 & 78.4 & 61.7 & 65.3 & 77.4 & 74.7 \\
        PSPNet~\cite{zhao2017pyramid} & 98.1 & \textbf{85.0} & 92.4 & 56.7 & 60.4 & 63.1 & 72.1 & 78.9 & 92.5 & 64.6 & 94.5 & 82.7 & 63.5 & 95.3 & 83.2 & 85.8 & 72.4 & 67.4 & 78.3 & 78.3 \\
        DeepLabv3~\cite{chen2017rethinking} & 98.0 & 84.4 & 92.3 & 52.9 & 59.5 & 62.6 & 70.6 & 78.1 & 92.4 & 65.0 & 94.5 & 82.2 & 63.0 & 95.3 & 83.3 & 87.1 & 71.5 & \textbf{70.3} & 77.5 & 77.9 \\
        DeepLabv3+~\cite{chen2018encoder} & \textbf{98.2} & 84.9 & 92.7 & \textbf{57.3} & 62.1 & 65.2 & 68.6 & 78.9 & 92.7 & 63.5 & \textbf{95.3} & 82.3 & 62.8 & 95.4 & \textbf{85.3} & 89.1 & 80.9 & 64.6 & 77.3 & 78.8 \\
    \midrule
        CANet(ours) & 98.0 & 84.2 & \textbf{92.7} & 51.3 & \textbf{62.7} & \textbf{67.8} & \textbf{73.6} & \textbf{81.4} & \textbf{92.8} & \textbf{65.0} & 95.2 & \textbf{84.4} & \textbf{67.4} & \textbf{95.7} & 84.1 & \textbf{91.8} & \textbf{84.0} & 68.9 & \textbf{79.7} & \textbf{80.0} \\
    \bottomrule
    \end{tabular}}
\end{center}
\label{table:citysval}
\end{table*}

It is noticed that CANet achieves no significant performance improvement than the DeepLabv3~\cite{chen2017rethinking} method (mIoU 78.68\% v.s. 78.45\%). We assume the reason lies in small scale changes exhibited in the dataset. To further validate the effectiveness, we conduct comparisons on the Cityscapes dataset~\cite{Cordts_2016_CVPR} of complicated scenes exhibiting massive scale changes. Table~\ref{table:citysval} reports the results, where DeepLabv3+~\cite{chen2018encoder} is based on Xception65~\cite{chollet2017xception} and others dilated ResNet101~\cite{he2016deep}. Even compared to the DeepLabv3+ model, CANet achieves significant performance improvements and gains the best in 13 out of 19 semantic categories. Qualitative improvements are illustrated in Fig.~\ref{fig:citysval}, where PSPNet~\cite{zhao2017pyramid} introduces inconsistent sub-region prediction because of large pooling stride and loss of spatial details, and DeepLabv3~\cite{chen2017rethinking} adopts dilated convolutions but exacerbates the gridding effect~\cite{wang2018understanding}. Differently, CANet employs encoder-decoder information flows with appropriate down-sampling scales, which constructs sufficient context features and brings adequate localization information. 

Note that we do not use FSM, decoder and auxiliary loss for the moment for fair comparison and to better address the basic series-parallel hybrid design of CAM.

\begin{table}
\caption{Ablation results of key components and auxiliary loss (AL)}
\begin{center}
\begin{tabular}{cccc|c}
\toprule
\textbf{CAM} & \textbf{FSM} & \textbf{Decoder} & \textbf{AL} & \textbf{mIoU(\%)} \\
\midrule\midrule
~ & ~ & ~ & ~ & 69.97 \\
~ & ~ & ~ & \checkmark & 70.48 \\
\midrule
\checkmark(w/o GF) & ~ & ~ & ~ & 78.53 \\
\checkmark(w/ GF) & ~ & ~ & ~ & \textbf{78.68} \\
\midrule
\checkmark & \checkmark & ~ & ~ & 78.90 \\
\checkmark & \checkmark & $1\times1$ Conv & ~ & 79.02 \\
\checkmark & \checkmark & $3\times3$ Conv & ~ & 79.48 \\
\checkmark & \checkmark & $3\times3$ Conv &  \checkmark & \textbf{79.69} \\
\bottomrule
\end{tabular}
\end{center}
\label{table:key_components}
\end{table}

\subsubsection{Ablation for Key Components}
\begin{figure}[t]
\begin{center}
\includegraphics[width=1.0\linewidth]{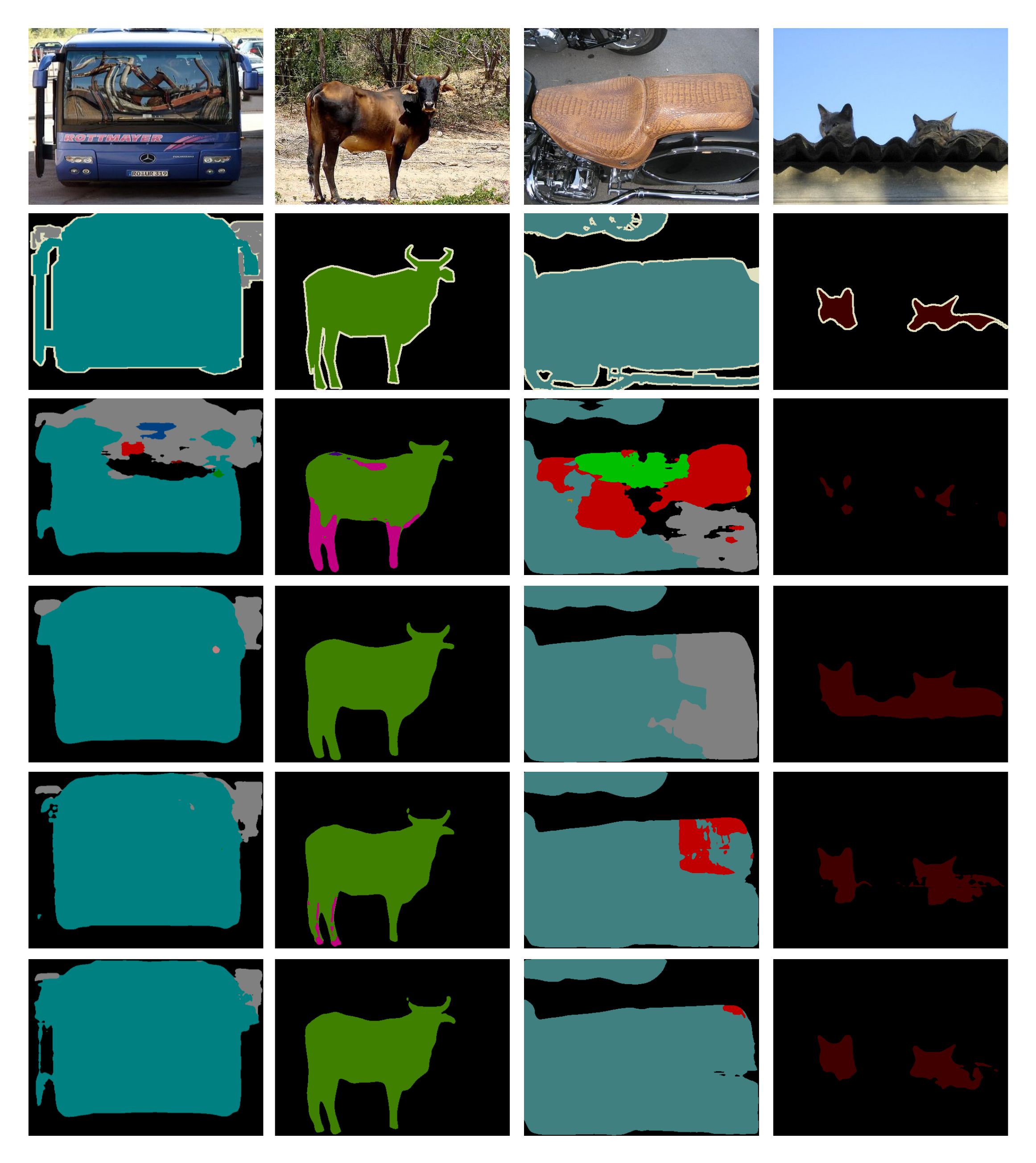}
\end{center}
\caption{Some visualized comparisons of different components on Pascal VOC 2012 validation set. \textbf{First row}: input images. \textbf{Second row}: ground truth. \textbf{Third row}: predictions of dilated FCN (baseline). \textbf{Fourth row}: predictions of CANet without FSM. \textbf{Fifth row}: predictions of CANet with $1\times1$ convolution based decoder. \textbf{Sixth row}: predictions of CANet with $3\times3$ convolution based decoder.}
\label{fig:pascal}
\end{figure}

We show the effectiveness of key components in CANet by adding them to the baseline one by one, and validate the benefits of GF in rejecting local ambiguities. The uppermost CF takes only the shared features as input when there is no GF. The experimental results are listed in Table~\ref{table:key_components}. We can grasp that: (1) CAM significantly improves the semantic segmentation performance, from mIoU 69.97\% to 78.68\% due to its powerful modeling ability of multi-scale contexts. (2) As assistant roles, GF obtains a global receptive field and FSM boosts feature re-fusion at channel domain. Both proffer minor enhancement to model performance. (3) The decoder brings spatial details of the low-level, thus refines the results. Moreover, the utilization of $3\times3$ convolution enhances semantic representations of both small and large objects of low-level features, thus achieves a double effect than the original $1\times1$ convolution. Fig.~\ref{fig:pascal} presents some visualized interpretations. It can be noticed that the decoder based on $3\times3$ convolution interjects less turbulence when recovering spatial resolution. (4) For both baseline and CANet, the use of auxiliary loss is beneficial to a better backbone feature extractor. We set the auxiliary loss weight $\lambda=0.1$ to gain better performance, as shown in Fig.~\ref{fig:comparisons}(b).

\begin{table}
\caption{Validation of improvement strategies}
\begin{center}
\begin{tabular}{cccc|c}
\toprule
\textbf{Backbone} & \textbf{MS-COCO} & \textbf{FT} & \textbf{MS/Flip} & \textbf{mIoU(\%)} \\
\midrule\midrule
ResNet50 & ~ & ~ & ~ & 79.69 \\
ResNet50 & \checkmark & ~ & ~ & 82.75 \\
ResNet50 & \checkmark & \checkmark & ~ & 83.45 \\
ResNet50 & \checkmark & \checkmark & \checkmark & 84.32 \\
ResNet101 & \checkmark & \checkmark & \checkmark & \textbf{84.90} \\
\bottomrule
\end{tabular}
\end{center}
\label{table:strategies}
\end{table}

\begin{table*}[]
\caption{Per-class scores on Pascal VOC 2012 test set}
\begin{center}
    \resizebox{1.0\linewidth}{!}{
    \begin{tabular}{l|cccccccccccccccccccc|c}
    \toprule
        Method & aero & bike & bird & boat & bottle & bus & car & cat & chair & cow & table & dog & horse & mbike & person & plant & sheep & sofa & train & tv & mIoU(\%) \\
    \midrule\midrule
        FCN~\cite{long2015fully} & 76.8 & 34.2 & 68.9 & 49.4 & 60.3 & 75.3 & 74.7 & 77.6 & 21.4 & 62.5 & 46.8 & 71.8 & 63.9 & 76.5 & 73.9 & 45.2 & 72.4 & 37.4 & 70.9 & 55.1 & 62.2 \\
        DeepLabv2~\cite{chen2017deeplab} & 84.4 & 54.5 & 81.5 & 63.6 & 65.9 & 85.1 & 79.1 & 83.4 & 30.7 & 74.1 & 59.8 & 79.0 & 76.1 & 83.2 & 80.8 & 59.7 & 82.2 & 50.4 & 73.1 & 63.7 & 71.6 \\
        DeconvNet~\cite{noh2015learning} & 89.9 & 39.9 & 79.7 & 63.9 & 68.2 & 87.4 & 81.2 & 86.1 & 28.5 & 77.0 & 62.0 & 79.0 & 80.3 & 83.6 & 80.2 & 58.8 & 83.4 & 54.3 & 80.7 & 65.0 & 72.5 \\
        DPN~\cite{liu2015semantic} & 87.7 & 59.4 & 78.4 & 64.9 & 70.3 & 89.3 & 83.5 & 86.1 & 31.7 & 79.9 & 62.6 & 81.9 & 80.0 & 83.5 & 82.3 & 60.5 & 83.2 & 53.4 & 77.9 & 65.0 & 74.1 \\
        Piecewise~\cite{lin2016efficient} & 90.6 & 37.6 & 80.0 & 67.8 & 74.4 & 92.0 & 85.2 & 86.2 & 39.1 & 81.2 & 58.9 & 83.8 & 83.9 & 84.3 & 84.8 & 62.1 & 83.2 & 58.2 & 80.8 & 72.3 & 75.3 \\
        ResNet38~\cite{wu2019wider} & 94.4 & 72.9 & 94.9 & 68.8 & 78.4 & 90.6 & 90.0 & 92.1 & 40.1 & 90.4 & 71.7 & 89.9 & 93.7 & \textbf{91.0} & 89.1 & 71.3 & 90.7 & 61.3 & 87.7 & 78.1 & 82.5 \\
        PSPNet~\cite{zhao2017pyramid} & 91.8 & 71.9 & 94.7 & 71.2 & 75.8 & 95.2 & 89.9 & 95.9 & 39.3 & 90.7 & 71.7 & 90.5 & 94.5 & 88.8 & 89.6 & 72.8 & 89.6 & 64.0 & 85.1 & 76.3 & 82.6 \\
        EncNet~\cite{zhang2018context} & 94.1 & 69.2 & \textbf{96.3} & \textbf{76.7} & \textbf{86.2} & 96.3 & 90.7 & 94.2 & 38.8 & 90.7 & 73.3 & 90.0 & 92.5 & 88.8 & 87.9 & 68.7 & 92.6 & 59.0 & 86.4 & 73.4 & 82.9 \\
        SDN~\cite{fu2019stacked} & \textbf{96.2} & 73.9 & 94.0 & 74.1 & 76.1 & 96.7 & 89.9 & \textbf{96.2} & \textbf{44.1} & 92.6 & 72.3 & 91.2 & 94.1 & 89.2 & 89.7 & 71.2 & 93.0 & 59.0 & 88.4 & 76.5 & 83.5 \\
        SeENet~\cite{pang2019towards} & 93.7 & 73.7 & 94.4 & 67.8 & 82.4 & 94.5 & \textbf{90.7} & 94.1 & 42.4 & 92.5 & 72.1 & 90.8 & 92.6 & 88.3 & 89.4 & 76.6 & 92.9 & \textbf{68.1} & 88.5 & 77.2 & 83.8 \\
        CFNet~\cite{zhang2019co} & 95.7 & 71.9 & 95.0 & 76.3 & 82.8 & 94.8 & 90.0 & 95.9 & 37.1 & 92.6 & 73.0 & \textbf{93.4} & 94.6 & 89.6 & 88.4 & 74.9 & \textbf{95.2} & 63.2 & 89.7 & 78.2 & 84.2 \\
        APCNet~\cite{he2019adaptive} & 95.8 & \textbf{75.8} & 84.5 & 76.0 & 80.6 & \textbf{96.9} & 90.0 & 96.0 & 42.0 & \textbf{93.7} & \textbf{75.4} & 91.6 & 95.0 & 90.5 & 89.3 & 75.8 & 92.8 & 61.9 & 88.9 & \textbf{79.6} & 84.2 \\
    \midrule
        CANet(ours) & 96.0 & 75.4 & 95.3 & 72.0 & 78.6 & 96.3 & 89.8 & 95.5 & 41.5 & 93.0 & 72.3 & 92.3 & \textbf{95.2} & 90.5 & \textbf{90.3} & \textbf{77.1} & 92.3 & 62.4 & \textbf{90.2} & 79.3 & \textbf{84.4}\\
    \bottomrule
    \end{tabular}}
\end{center}
\label{table:pascaltest}
\end{table*}

\subsubsection{Improvement Strategies}In this section, we evaluate several improvement strategies that further improve the segmentation performance of CANet in Table~\ref{table:strategies}, including (1) MS-COCO: pre-training with MS-COCO~\cite{lin2014microsoft} dataset, (2) FT: fine-tune on the original dataset, (3) MS/Flip: testing with multi-scale inputs as well as their left-right mirrors, and (4) employing a deeper backbone network.

\subsubsection{Comparisons with State-of-the-Arts}
We conduct experiments on the test set to compare with other prior methods based on ImageNet pre-trained ResNet101. CANet is first pre-trained on the augmented dataset, then fine-tuned with the original \emph{trainval} images. Results based on multi-scale and flipping testing skills are reported in Table~\ref{table:pascaltest}. CANet achieves mIoU 84.4\% and outperforms all other approaches. To be specific, CANet surpasses the SDN~\cite{fu2019stacked} model of purely in-series encoder-decoder by 0.9 points mIoU with simplicity, which demonstrates the superiority of the series-parallel hybrid design. Additionally, when using MS-COCO~\cite{lin2014microsoft} pre-training, we obtain mIoU 87.2\%.

\begin{table}
\caption{Results on Pascal Context dataset. Note that we report mIoU on 60 classes with background}
\begin{center}
    \begin{tabular}{lc|c}
    \toprule
        \textbf{Method} & \textbf{Backbone)} & \textbf{mIoU(\%)} \\
    \midrule\midrule
        FCN-8s~\cite{long2015fully} & - & 37.8 \\
        Piecewise~\cite{lin2016efficient} & - & 43.4 \\
        DeepLabv2~\cite{chen2017deeplab} & ResNet101 & 45.7 \\
        RefineNet~\cite{lin2017refinenet} & ResNet152 & 47.3 \\
        PSPNet~\cite{zhao2017pyramid} & ResNet101 & 47.8 \\
        EncNet~\cite{zhang2018context} & ResNet101 & 51.7 \\
        DANet~\cite{fu2019dual} & ResNet101 & 52.6 \\
        ANL~\cite{zhu2019asymmetric} & ResNet101 & 52.8 \\
        BFP~\cite{ding2019boundary} & ResNet101 & 53.6 \\
        CPNet~\cite{yu2020context} & ResNet101 & 53.9 \\
    \midrule
        CANet(ours) & ResNet101 & \textbf{54.3} \\
    \bottomrule
    \end{tabular}
\end{center}
\label{table:pcontext}
\end{table}

\subsection{Results on Pascal Context}
Pascal Context~\cite{mottaghi2014role} is a scene parsing dataset, containing 4998 images for training and 5105 images for validation. Following prior works~\cite{zhang2018context,yu2020context}, we use the most common 59 categories for this benchmark and consider all the other classes as background. Results in Table~\ref{table:pcontext} demonstrate CANet achieves state-of-the-art performance. Compared with DANet~\cite{fu2019dual} and ANL~\cite{zhu2019asymmetric} that focus on feature fusion via applying non-local operations on pixel pairs, this paper pays more attention to the feature propagation paradigm and achieves significant performance improvements of 1.7 points and 1.5 points mIoU, respectively.

\begin{table}
\caption{Experimental results on Cityscapes test set. $\dagger$ here means employing validation images for training}
\begin{center}
    \begin{tabular}{lc|c}
    \toprule
        \textbf{Method} & \textbf{Backbone)} & \textbf{mIoU(\%)} \\
    \midrule\midrule
        DeepLabv2~\cite{chen2017deeplab} & ResNet101 & 70.4 \\
        RefineNet$\dagger$~\cite{lin2017refinenet} & ResNet101 & 73.6 \\
        ResNet38~\cite{wu2019wider} & WiderResNet38 & 78.4 \\
        PSPNet~\cite{zhao2017pyramid} & ResNet101 & 78.4 \\
        DFN$\dagger$~\cite{yu2018learning}  & ResNet101 & 79.3 \\
        PSANet$\dagger$~\cite{zhao2018psanet} & ResNet101 & 80.1 \\
        DenseASPP$\dagger$~\cite{yang2018denseaspp} & DenseNet161 & 80.6 \\
        SeENet$\dagger$~\cite{pang2019towards} & ResNet101 & 81.2 \\
        ANL$\dagger$~\cite{zhu2019asymmetric} & ResNet101 & 81.3 \\
        DANet$\dagger$~\cite{fu2019dual} & ResNet101 & 81.5 \\
        ACFNet$\dagger$~\cite{zhang2019acfnet} & ResNet101 & 81.8 \\
        SPNet$\dagger$~\cite{hou2020strip} & ResNet101 & 82.0 \\
    \midrule
        CANet(ours)$\dagger$ & ResNet101 & \textbf{82.6} \\
    \bottomrule
    \end{tabular}
\end{center}
\label{table:citys}
\end{table}

\subsection{Results on Cityscapes}
Cityscapes~\cite{Cordts_2016_CVPR} describes complex street scenes with high-resolution images that are fine labeled in 19 classes for semantic segmentation. Following the standard settings, we do not use coarse labeled data for evaluation, resulting in 2975 images for training, 500 images for validation and 1525 images for testing. We report quantitative comparisons in Table~\ref{table:citys}. Results confirm that CANet can tackle complex scale variations and obtain promising performance for high-resolution street scene parsing. More specifically, the series-parallel hybrid structure of CANet has similarities with DenseASPP~\cite{yang2018denseaspp} that we can obtain the former by removing some dense connections in the latter. The key difference lies in the information flow where CANet adopts a shallow encoder-decoder with suitable down-sampling scales and DenseASPP single dilated $3\times3$ convolution. Therefore, CANet can integrate more localization information, achieving a significant performance improvement of 2.0 points mIoU. We note that DenseASPP employs a more powerful backbone network of DenseNet~\cite{huang2017densely}.

\subsection{Results on CamVid}
CamVid~\cite{brostow2009semantic} is a street scene dataset that contains both light and dark conditions. We use the dataset described by SegNet~\cite{badrinarayanan2017segnet} that contains 367 images for training, 100 images for validation, and 233 images for testing, all labeled with 11 semantic categories. We train CANet with training and validation images and report the performance on the testing images. Table~\ref{table:camvid} reports the results. The proposed CANet is significantly better than existing methods which firmly evidence the superiority. Fig.~\ref{fig:camvid} gives some visualizations. Because of the capability to capture multi-scale contextual information, CANet is able to ameliorate poor object delineation and small spurious regions.
\begin{table}
\caption{Experimental results on CamVid dataset (11 classes). $\dagger$ here means evaluation with $720\times 960$ images}
\begin{center}
    \begin{tabular}{l|cc}
    \toprule
        \textbf{Method} & \textbf{PA(\%)} & \textbf{mIoU(\%)} \\
    \midrule\midrule
        SegNet~\cite{badrinarayanan2017segnet} & 62.5 & 46.4 \\
        DeconvNet~\cite{noh2015learning} & 85.6 & 48.9 \\
        Bayesian SegNet~\cite{kendall2015bayesian} & 86.9 & 63.1 \\
        Dilation8~\cite{yu2015multi} & 79.0 & 65.3 \\
        HDCNN-448+TL~\cite{wang2017hierarchically} & 90.9 & 65.6 \\
        Dilation8+FSO~\cite{kundu2016feature} & 88.3 & 66.1 \\
        FC-DenseNet103~\cite{jegou2017one} & 91.5 & 66.9 \\
        DCDN~\cite{fu2017densely} & 91.4 & 68.4 \\
        SeENet~\cite{pang2019towards} & - & 68.4 \\
        SDN~\cite{fu2019stacked} & 91.7 & 69.6 \\
        BFP~\cite{ding2019boundary} & - & 74.1 \\
    \midrule
        CANet(ours) & \textbf{93.9} & \textbf{75.6} \\
        CANet(ours)$\dagger$ & \textbf{94.4} & \textbf{78.6} \\
    \bottomrule
    \end{tabular}
\end{center}
\label{table:camvid}
\end{table}

\subsection{Results on SUN-RGBD}
SUN-RGBD~\cite{song2015sun} dataset has a total of 10335 indoor images, of which 5280 images are for training and 5050 images for testing. It provides pixel-wise labeling for 37 semantic labels. There are various objects in one image scene and they differ in shapes, sizes and even spatial poses, which makes SUN-RGBD one of the most challenging datasets. In this paper, we only utilize RGB modality for experiments. Quantitative results reported in Table~\ref{table:sunrgbd} demonstrate that our CANet achieves optimal performance.

\begin{table}
\caption{Quantitative results on SUN-RGBD dataset (37 classes) which only use RGB modality for evaluation}
\begin{center}
    \begin{tabular}{l|cc}
    \toprule
        \textbf{Method} & \textbf{PA(\%)} & \textbf{mIoU(\%)} \\
    \midrule\midrule
        FCN~\cite{long2015fully} & 68.2 & 27.4 \\
        DeconvNet~\cite{noh2015learning} & 66.1 & 22.6 \\
        SegNet~\cite{badrinarayanan2017segnet} & 72.6 & 31.8 \\
        Bayesian SegNet~\cite{kendall2015bayesian} & 71.2 & 30.7 \\
        DeepLabv2~\cite{chen2017deeplab} & 71.9 & 32.1 \\
        Piecewise~\cite{lin2016efficient} & 78.4 & 42.3 \\
        RefineNet~\cite{lin2017refinenet} & 80.4 & 45.7 \\
        Ding et al.~\cite{ding2018context} & 81.4 & 47.1 \\
    \midrule
        CANet(ours) & \textbf{82.0} & \textbf{48.3} \\
    \bottomrule
    \end{tabular}
\end{center}
\label{table:sunrgbd}
\end{table}

\subsection{Results on GATECH}
We further evaluate CANet on the GATECH dataset~\cite{hussain2013geometric} that has somewhat noisy annotations. GATECH is a large outdoor scene video dataset, which contains 8 semantic annotation categories, including 63 videos with a total of 12241 frames for training and 38 videos with a total of 7071 frames for testing. Due to the vast redundancy between video frames, similar to S. J{\'e}gou et al.~\cite{jegou2017one}, we extract images every five frames for training and use all test frames to evaluate the model performance. Table~\ref{table:gatech} reports the results. Without the use of temporal information, our CANet surpasses existing methods with a large margin, proving that CANet obtains robust context features even with noisy annotations.

\begin{table}
\caption{Results on GATECH dataset}
\begin{center}
    \resizebox{1.0\linewidth}{!}{
    \begin{tabular}{lc|cc}
    \toprule
        \textbf{Method} & \textbf{Temporal Info} & \textbf{PA(\%)} & \textbf{mIoU(\%)} \\
    \midrule\midrule
        3D-V2V-scratch~\cite{tran2016deep} & Yes & 66.7 & - \\
        3D-V2V-finetune~\cite{tran2016deep} & Yes & 76.0 & - \\
        FC-DenseNet103~\cite{jegou2017one} & No & 79.4 & - \\
        HDCNN-448+TL~\cite{wang2017hierarchically} & Yes & 82.1 & 48.2 \\
        DCDN~\cite{fu2017densely} & No & 83.5 & 49.0 \\
        SDN~\cite{fu2019stacked} & No & 84.6 & 53.5 \\
    \midrule
        CANet(ours) & No & \textbf{86.6} & \textbf{56.0} \\
    \bottomrule
    \end{tabular}}
\end{center}
\label{table:gatech}
\end{table}

\section{Conclusion}
\label{section:conclusion}
This paper proposes the Chained Context Aggregation Module to enrich multi-scale contexts from a novel spatial perspective. With the fundamental series-parallel hybrid design of information flows, CAM effectively encodes diverse region-based contexts through the naturally developed pre-fusion and re-fusion process, giving a remarkable improvement on performance. The series-parallel hybrid structure not only enhances feature interaction but also is friendly to model training. As an extension work, we adopt $3\times3$ convolution to refine the decoder, which introduces less turbulence when recovering spatial resolution. Extensive experiments on six challenging datasets indicate the effectiveness and advancement of the proposed CANet. 

However, we still use summation and concatenation for feature fusion apart from the simple attention model FSM. Attention mechanisms show excellent abilities for modeling pixel-wise and region-wise dependencies, taking semantic segmentation a big step forward. We expect to do more in-depth work in multi-scale feature fusion based on the proposed feature encoding paradigm, bringing new vitality to the semantic segmentation community.


%




\ifCLASSOPTIONcaptionsoff
  \newpage
\fi



\bibliographystyle{IEEEtran}
\bibliography{IEEEabrv,ref}
\end{document}